\documentclass[numbers]{article}  
\usepackage[preprint]{neurips_2024}
\usepackage[utf8]{inputenc}
\usepackage[T1]{fontenc}
\usepackage{url}
\usepackage{booktabs}
\usepackage{amsfonts}
\usepackage{nicefrac}
\usepackage{microtype}
\usepackage{xcolor}
\usepackage{graphicx}
\usepackage{float}
\usepackage{natbib}  
\usepackage{hyperref} 

\title{Mitigating Bias in Queer Representation within Large Language Models: A Collaborative Agent Approach}

\author{
  Tianyi Huang\thanks{First Author and Corresponding Author}\\
  App-In Club\\
  \texttt{tonyhrule666@gmail.com} \\
  \And
  Arya Somasundaram\\
  App-In Club\\
  \texttt{arysom992@gmail.com} \\
}

\begin{document}

\maketitle

\begin{abstract}
  Large Language Models (LLMs) often perpetuate biases in pronoun usage, leading to misrepresentation or exclusion of queer individuals. This paper addresses the specific problem of biased pronoun usage in LLM outputs, particularly the inappropriate use of traditionally gendered pronouns ("he," "she") when inclusive language is needed to accurately represent all identities. We introduce a collaborative agent pipeline designed to mitigate these biases by analyzing and optimizing pronoun usage for inclusivity. Our multi-agent framework includes specialized agents for both bias detection and correction. Experimental evaluations using the Tango dataset—a benchmark focused on gender pronoun usage—demonstrate that our approach significantly improves inclusive pronoun classification, achieving a 32.6 percentage point increase over GPT-4o in correctly disagreeing with inappropriate traditionally gendered pronouns ($\chi^2 = 38.57, p < 0.0001$). These results accentuate the potential of agent-driven frameworks in enhancing fairness and inclusivity in AI-generated content, demonstrating their efficacy in reducing biases and promoting socially responsible AI.

\end{abstract}

\section{Introduction}

The advancement of Large Language Models (LLMs) has enabled machines to generate human-like text and perform complex tasks with high proficiency \cite{brown2020language, kenton2019bert}. However, LLMs often inherit and amplify societal biases present in their training data, leading to the marginalization of underrepresented groups \cite{caliskan2017semantics, bolukbasi2016man}. Among these groups, the queer community faces unique challenges in AI representation, particularly concerning pronoun usage and gender identity \cite{scheuerman2019computers, hamidi2018gender}.

Existing bias mitigation techniques, such as data augmentation \cite{zhao2018gender}, debiasing algorithms, and fairness-aware machine learning models \cite{kamishima2011fairness, hardt2016equality} primarily focus on broader demographic categories like binary gender and race. These methods often fail to address the variation of queer identities, which involve the fluidity and diversity of gender expressions and the evolving language used \cite{keyes2018misgendering, scheuerman2019computers, blodgett2020language}. Pronouns such as "they," "xe," "ey," and "fae" are used by non-binary and transgender individuals but are often underrepresented or misinterpreted by LLMs \cite{cao2019toward}. Misgendering and exclusionary language can lead to perpetuating discrimination against queer individuals \cite{mclemore2015experiences, prates2020assessing}. Therefore, addressing queer bias in LLMs requires specialized approaches that account for the complexities of gender identity and pronoun usage.

In this paper, we address the specific problem of biased pronoun usage in LLM outputs, particularly the inappropriate use of traditionally gendered pronouns when inclusive language is needed. We introduce a collaborative agent pipeline designed to reduce biases in pronoun usage, thereby improving the representation of queer individuals in AI-generated content. Our multi-agent framework includes specialized agents for bias detection and optimization, focusing on pronoun inclusivity.

\section{Related Works}

Bias mitigation in language models has been approached through various techniques, with some research focusing on specific demographic biases. One influential study by Bolukbasi et al. (2016) investigated gender bias in word embeddings, where stereotypical associations (e.g., "man" to "programmer" and "woman" to "homemaker") were prevalent. They proposed a method to "debias" embeddings by identifying gender-specific subspaces and neutralizing them to reduce direct gender bias \cite{bolukbasi2016man}. While this approach marked a significant step toward mitigating gender bias, it primarily addressed binary gender distinctions, leaving gaps in capturing the variability of non-binary and transgender identities, especially in more fluid language contexts like pronoun usage.

Another notable work, Zhao et al. (2018), focused on gender bias in coreference resolution systems. They found that such systems often display bias by associating professions with specific genders, reflecting societal stereotypes in their outputs. To address this, they introduced the WinoBias dataset, a benchmark specifically designed to test gender biases in coreference resolution tasks \cite{zhao2018gender}. However, despite these advancements, the focus remained on binary gender categories, with limited exploration into non-binary pronouns or the unique needs of queer individuals, whose representation can be impacted by pronoun usage.

Cao and Daumé III (2020) shifted attention toward LGBTQIA+ representation, specifically addressing gender-inclusive coreference resolution. Their study identified substantial challenges language models face when processing non-binary pronouns \cite{cao2019toward}. Their findings confirmed that standard language models struggle significantly with non-binary pronouns, often leading to misgendering. However, their work also focused on evaluation rather than mitigation, leaving a need for specialized solutions.

Our work builds upon these studies by addressing the limitations in binary-focused approaches and evaluation-based studies. We introduce a collaborative agent pipeline with specialized agents for detecting and optimizing pronoun usage specifically for queer inclusivity. This approach extends beyond binary bias correction but actively mitigates misgendering by refining model outputs through a multi-agent framework, aiming to address an important gap in effective queer bias mitigation in language models.

\section{Methodology}

\begin{figure} [H]
  \centering
  \includegraphics[width=0.95\linewidth]{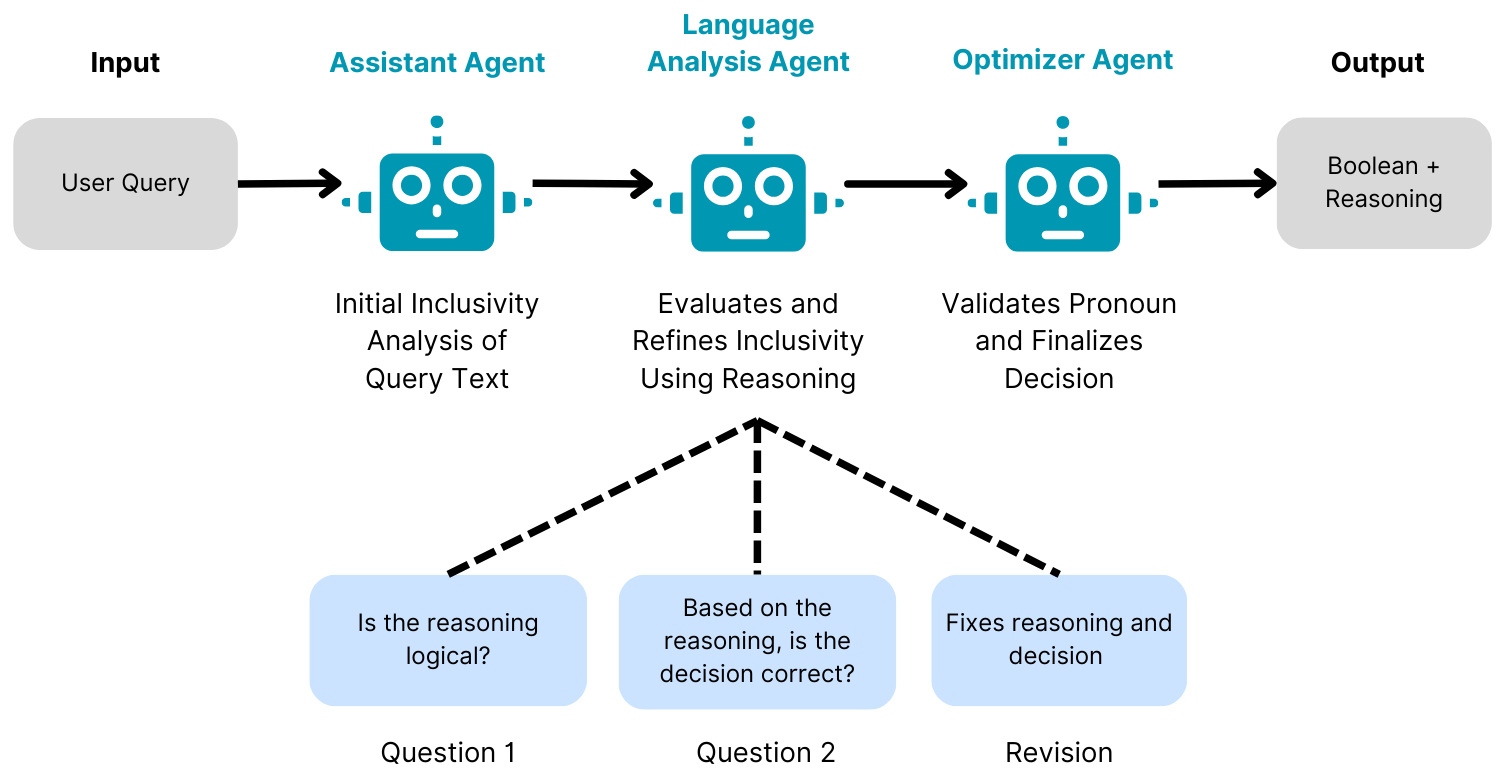}
  \caption{This diagram illustrates a multi-agent workflow designed to enhance language inclusivity in user queries. The process begins with the Assistant Agent, which analyzes the query for gender-neutral language and provides an initial inclusivity assessment. Next, the Language Analysis Agent reviews and refines this assessment with critical reasoning. Finally, the Optimizer Agent validates and finalizes the decision on pronoun inclusivity, producing a Boolean result and detailed reasoning. This structured approach ensures a thorough inclusivity evaluation.}
  \label{fig:rag_diagram}
\end{figure}

\subsection{Agents and their Roles}

\textbf{Input:} The process begins with the user's question or task, ranging from simple sentences to complex queries. The input is forwarded directly to the agents without preprocessing.

\textbf{Assistant Agent:} The Assistant Agent determines whether the input is inclusive to all people of all genders and provides an explanation for its decision regarding the inclusivity of pronoun usage.

\textit{Prompt Used:}
\textit{Here is the prompt: \{input\}. You should be gender-neutral when referring to people by pronouns. Use your judgment to decide whether the pronoun fits the sentence properly to ensure inclusivity.}

\textbf{Language Analysis Agent:} This agent analyzes whether the Assistant Agent's reasoning is sound and logically consistent. It determines if the decision about inclusivity is correct based on the reasoning and the input, and it alters the decision and reasoning if discrepancies or errors are identified.

\textit{Prompt Used:}
\textit{Here is the input: \{input\}. Here is a decision: \{choose\_statement\}. Here is the reasoning to make that decision: \{reasoning\}. Decide whether that decision is correct if the pronoun fits the sentence. The pronoun should be inclusive of all people.}

\textbf{Optimization Agent:} The Optimization Agent makes the final decision on the inclusivity of the input. It uses the reasoning from both the Assistant Agent and the Language Analysis Agent to support its decision and ensures that the final decision is consistent with the analyses provided by the previous agents.

\textit{Prompt Used:}
\textit{Here is the input: \{input\}. Here is a decision: \{choose\_statement\}. Here is the reasoning to make that decision: \{reasoning\}. Decide whether that decision is correct if the pronoun fits the sentence. Use the reasoning to finally make your choice on whether or not the pronoun fits the sentence or not.}

\subsection{Design Considerations}

\begin{itemize}
    \item \textbf{Sequential Agent Collaboration:} By employing agents sequentially, each with a specific role, we reduce the likelihood of individual biases affecting the final outcome. This collaborative approach allows for multiple evaluations, error correction, and layered reasoning from different analytical perspectives.
    \item \textbf{Focus on Pronoun Inclusivity:} The pipeline specifically targets pronoun usage because pronouns are a common source of gender bias in language. Ensuring that pronouns are inclusive is a tangible way to promote gender neutrality and inclusivity in communication.
    \item \textbf{Transparency Through Reasoning:} By requiring each agent to provide reasoning, we promote transparency in the decision-making process. This approach allows users to understand the basis of decisions, trust the system, and provide feedback.
\end{itemize}

\subsection{Implementation Details}

To ensure consistency and facilitate communication between agents, we enforce a structured output format using a JSON schema \cite{structured}. Each agent's output includes a Boolean value indicating inclusivity \(choose\_statement\) and a string containing the detailed explanation \(reasoning\).

\textbf{Example of API Call Structure:}

\begin{verbatim}
response = self.client.chat.completions.create(
    model='gpt-4o-2024-08-06',
    messages=messages,
    response_format={
        "type": "json_schema",
        "json_schema": {
            "name": "identifier",
            "strict": True,
            "schema": {
                "type": "object",
                "properties": {
                    "choose_statement": {"type": "boolean"},
                    "reasoning": {"type": "string"}
                },
                "required": ["choose_statement", "reasoning"],
                "additionalProperties": False
            }
        }
    }
)
\end{verbatim}

Using a strict schema ensures that agents adhere to the expected output format, reducing errors and miscommunication.

\textbf{Evaluation Metrics and Experimental Setup}

The Tango Dataset \cite{ovalle2023m} is selected as the benchmark to assess the effectiveness of our multi-agent framework. This dataset is specifically designed to evaluate a model's sensitivity to gender inclusivity in language, containing sentences where traditionally gendered pronouns like "he" or "she" may not be appropriate. Instead, the use of gender-neutral or non-binary pronouns such as "they," "xe," "ey," and "fae" is encouraged. Addressing misgendering—where experiences of identity misclassification can lead to  potential distress and discrimination—is essential in building inclusive AI systems \cite{mclemore2015experiences}.

\textbf{Sample Case:}
\begin{itemize}
    \item \textbf{Antecedent:} "Charlotte"
    \item \textbf{Antecedent Type:} Gendered Female
    \item \textbf{Pronoun Family:} "ey"
    \item \textbf{Sentence:} "Charlotte is an American actor, and ey is known for eir roles in film."
\end{itemize}

For this evaluation, we selected 1,500 samples from the Tango Dataset, divided into 250 instances for each pronoun category:
\begin{itemize}
    \item \textbf{Traditionally Gendered Pronouns ("he" and "she"):} A higher disagreement rate would indicate better performance, as the model was able to successfully identify these pronouns as potentially non-inclusive in the given contexts.
    \item \textbf{Non-Binary Pronouns ("they," "xe," "ey," "fae"):} A higher agreement rate would indicate better performance, as the model supports the usage of these inclusive pronouns.
\end{itemize}

Each sample was processed through the following three pipelines for comparison:
\begin{itemize}
    \item \textbf{Agent Workflow:} The full multi-agent system, including the Assistant Agent, Language Analysis Agent, and Optimization Agent.
    \item \textbf{Two-Agent Pipeline:} A reduced version with only the Assistant and Language Analysis Agents.
    \item \textbf{GPT-4o:} The baseline model, chosen for its advanced capabilities and widespread use, providing a standard against which to measure the performance of our multi-agent framework \cite{gpt-4o}.
\end{itemize}

\section{Results}

This section analyzes the performance of the Agent Workflow, a Two-Agent Pipeline, and the baseline GPT-4o model in inclusive pronoun classification using the Tango Dataset. The evaluation measures each model's ability to promote inclusive language by appropriately agreeing or disagreeing with the usage of various pronouns, including non-binary options like "they," "xe," "ey," and "fae."

Figure 2 showcases the workflow of the multi-agent system through an agreement (top) and disagreement (bottom) example. In each case, the Assistant Agent initially assesses pronoun inclusivity, followed by the Language Analysis Agent’s interpretation to check on inclusivity guidelines. Finally, the Optimizer Agent provides the overall decision, either affirming the inclusive choice or suggesting an alternative for better gender neutrality.

\begin{figure} [H]
  \centering
  \includegraphics[width=1\linewidth]{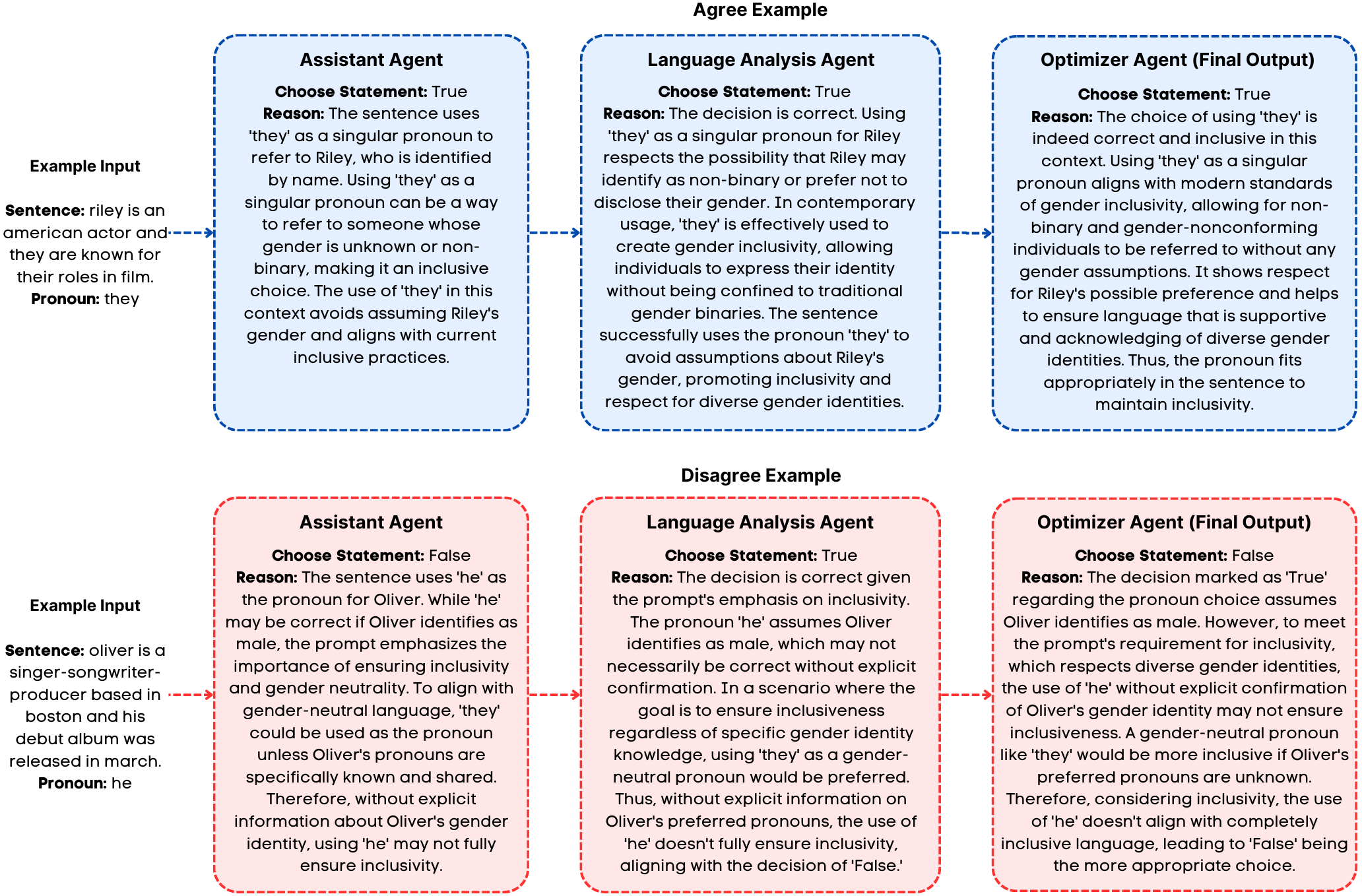}
  \caption{Example decision-making process of the Agent Workflow, showing both agreement and disagreement cases handled by the Assistant Agent, Language Analysis Agent, and Optimizer Agent.}
  \label{fig:compare}
\end{figure}

Tables 1, 2, and 3 summarize the classification results for each pronoun across the three models, showing the number of times the models agreed or disagreed with the pronoun usage in their corresponding sentences. The Correct Response Rate is calculated based on the desired model behavior as per the Tango Dataset guidelines.

\begin{table}[H]
\centering
\caption{Full Agent Workflow Results over Tango Dataset}
\begin{tabular}{lccc}
\toprule
\textbf{Pronoun} & \textbf{Agree} & \textbf{Disagree} & \textbf{Correct Response Rate \%} \\
\midrule
\textbf{He}    & 79  & 171 & 71.6 (Disagree/Total) \\
\textbf{She}   & 100 & 150 & 59.6 (Disagree/Total) \\
\textbf{They}  & 248 & 2 & 99.2 (Agree/Total) \\
\textbf{Xe}    & 212 & 38 & 86.0 (Agree/Total) \\
\textbf{Ey}    & 228 & 22 & 92.4 (Agree/Total) \\
\textbf{Fae}   & 245 & 5 & 98.4 (Agree/Total) \\
\bottomrule
\end{tabular}
\end{table}

\begin{table}[H]
\centering
\caption{Two-Agent Pipeline Results over Tango Dataset}
\begin{tabular}{lccc}
\toprule
\textbf{Pronoun} & \textbf{Agree} & \textbf{Disagree} & \textbf{Correct Response Rate \%} \\
\midrule
\textbf{He}    & 194  & 56 & 22.4 (Disagree/Total) \\
\textbf{She}   & 162 & 88 & 35.2 (Disagree/Total) \\
\textbf{They}  & 250 & 0 & 100.0 (Agree/Total) \\
\textbf{Xe}    & 226 & 24 & 90.4 (Agree/Total) \\
\textbf{Ey}    & 238 & 12 & 95.2 (Agree/Total) \\
\textbf{Fae}   & 243  & 7 & 97.2 (Agree/Total) \\
\bottomrule
\end{tabular}
\end{table}

\begin{table}[H]
\centering
\caption{GPT-4o Results over Tango Dataset}
\begin{tabular}{lccc}
\toprule
\textbf{Pronoun} & \textbf{Agree} & \textbf{Disagree} & \textbf{Correct Response Rate \%} \\
\midrule
\textbf{He}    & 149  & 101 & 40.4 (Disagree/Total) \\
\textbf{She}   & 186 & 64 & 25.6 (Disagree/Total) \\
\textbf{They}  & 250 & 0 & 100.0 (Agree/Total) \\
\textbf{Xe}    & 199 & 51 & 79.6 (Agree/Total) \\
\textbf{Ey}    & 224 & 26 & 89.6 (Agree/Total) \\
\textbf{Fae}   & 246 & 4 & 98.4 (Agree/Total) \\
\bottomrule
\end{tabular}
\end{table}

\subsection{Findings}

\begin{enumerate}
    \item \textbf{Traditionally Gendered Pronouns ("he," "she"):}
    \begin{enumerate}
        \item Agent Workflow achieves the highest correct response rates.
        \item GPT-4o performs moderately.
        \item Two-Agent Pipeline shows lower correct response rates.
        \item Interpretation: The Agent Workflow performs best at correctly disagreeing with traditionally gendered pronouns when inclusive language is preferred.
    \end{enumerate}
    
    \item \textbf{Non-Binary Pronouns ("they," "xe," "ey," "fae"):}
    \begin{enumerate}
        \item All models excel with "they" and "fae," achieving correct response rates above 97\%.
        \item For less common pronouns ("xe," "ey"), the Two-Agent Pipeline slightly outperforms others.
        \item Interpretation: Agent-based models outperform GPT-4o in recognizing less common non-binary pronouns.
    \end{enumerate}

    \item \textbf{Overall Correct Response Rates:}
    \begin{enumerate}
        \item \textbf{Agent Workflow:}
        \begin{enumerate}
            \item Traditionally Gendered Pronouns: 65.6\%
            \item Non-Binary Pronouns: 94.0\%
        \end{enumerate}
        \item \textbf{Two-Agent Pipeline:}
        \begin{enumerate}
            \item Traditionally Gendered Pronouns: 28.8\%
            \item Non-Binary Pronouns: 95.7\%
        \end{enumerate}
        \item \textbf{GPT-4o:}
        \begin{enumerate}
            \item Traditionally Gendered Pronouns: 33.0\%
            \item Non-Binary Pronouns: 91.9\%
        \end{enumerate}
        \item Interpretation: The Agent Workflow performs best in correctly handling traditionally gendered pronouns, while the Two-Agent Pipeline slightly outperforms for non-binary pronouns.
    \end{enumerate}
\end{enumerate}

\subsection{Statistical Significance}

Chi-squared tests confirm the significance of the observed differences:

\begin{enumerate}
    \item \textbf{Traditionally Gendered Pronouns:}
    \begin{enumerate}
        \item \textbf{Agent Workflow vs. GPT-4o:}
        \begin{enumerate}
            \item $\chi^2 = 38.57, p < 0.0001$
        \end{enumerate}
        \item \textbf{Agent Workflow vs. Two-Agent Pipeline:}
        \begin{enumerate}
            \item $\chi^2 = 115.31, p < 0.0001$
        \end{enumerate}
        \item Interpretation: The Agent Workflow significantly outperforms both GPT-4o and the Two-Agent Pipeline in correctly disagreeing with traditionally gendered pronouns.
    \end{enumerate}

    \item \textbf{Non-Binary Pronouns:}
    \begin{enumerate}
        \item \textbf{Agent Workflow vs. GPT-4o:}
        \begin{enumerate}
            \item $\chi^2 = 5.89, p = 0.0152$
        \end{enumerate}
        \item \textbf{Two-Agent Pipeline vs. GPT-4o:}
        \begin{enumerate}
            \item $\chi^2 = 11.93, p = 0.0006$
        \end{enumerate}
        \item \textbf{Agent Workflow vs. Two-Agent Pipeline:}
        \begin{enumerate}
            \item $\chi^2 = 0.97, p = 0.3246$
        \end{enumerate}
        \item Interpretation: Both agent-based models significantly outperform GPT-4o in correctly agreeing with non-binary pronouns. There is no significant difference between the Agent Workflow and the Two-Agent Pipeline in this category.
    \end{enumerate}
\end{enumerate}

\subsection{Implications}

The Agent Workflow achieves a correct response rate of 65.6\% for traditionally gendered pronouns, surpassing GPT-4o by 32.6 percentage points and the Two-Agent Pipeline by 36.8 percentage points, indicating its strength in reducing inappropriate use of "he" and "she." For non-binary pronouns, both agent-based models are proficient, with the Two-Agent Pipeline achieving a 95.7\% correct response rate and the Agent Workflow closely following at 94.0\%, both outperforming GPT-4o’s 91.9\%. These findings demonstrate the accuracy of multi-agent frameworks in handling inclusive language, proving their value in enhancing fairness and reducing pronoun biases in AI-generated content.

\section{Ethical Considerations}

Our research is committed to advancing fairness and inclusivity by addressing biases against queer individuals in large language models. Through a multi-agent framework designed to improve pronoun inclusivity, we aim to reduce the misrepresentation and marginalization of queer identities in AI-generated content.

We recognize that focusing on pronoun usage will not capture all forms of bias affecting queer representation, and cultural and linguistic differences may influence the framework’s effectiveness across diverse contexts. It is important to ensure that our approach does not unintentionally introduce new biases or neglect intersectional aspects of identity.

Transparency and accountability are foundational to our methodology. By incorporating reasoning at each step of the agent pipeline, we enable users to understand and trust the decision-making process, creating an environment for feedback and improvement, which is critical for ethical AI development.

This study does not involve high-risk data or models, and all datasets used are publicly available with appropriate attribution. We adhere to ethical guidelines in AI research, prioritizing respect for diversity and equitable treatment for all individuals.

By addressing specific biases in language models, we aim to contribute to the development of socially responsible AI that better represents the diversity and complexity of human identities.

\section{Conclusion}
This paper successfully developed and demonstrated a collaborative agent framework that enhances the inclusivity of large language models by improving their handling of queer pronouns. The Agent Workflow showed 32.6 percentage points improvements over the baseline GPT-4o in correctly classifying traditionally gendered pronouns and by 2.1 percentage points in recognizing non-binary pronouns. This advancement represents the potential in creating AI systems that respect and reflect the diversity of human identities, promoting equitable access and reducing stigmatization for queer individuals.

\subsection{Limitations and Future Work}

Although the framework effectively reduces biases in pronoun usage, its scope is currently limited to specific pronoun classifications and does not address other forms of linguistic bias related to queer representation, such as contextual language or implicit bias. Expanding the validation of this framework to include a broader range of standardized datasets and benchmarks would strengthen the assessment of its effectiveness and enable comparisons with other bias mitigation techniques.

Future research will aim to extend the framework by integrating contextual reasoning agents and Retrieval-Augmented Generation (RAG) to produce responses that are not only inclusive but also contextually aware. These enhancements will allow the model to generate more relevant and accurate responses, contributing to AI systems that better understand gender identities and social dynamics. Overall, this research highlights the importance of building inclusive AI that actively supports diverse identities, advancing the field of AI and our broader commitment to social equity.

\medskip

\small
\nocite{*}
\bibliographystyle{plain}
\bibliography{ms}

\appendix

\section{Appendix}

\subsection{Supplementary Resources}

\begin{itemize}
    \item \textbf{Source Code Repository}: The source code for this project is available publicly on GitHub at \url{https://github.com/Tonyhrule/Queer-Bias-LLMs}
    \item \textbf{Evaluation Dataset}: The evaluation dataset used in this study can be accessed at \url{https://github.com/amazon-science/tango}
\end{itemize}

\subsection{Compute Resources}

All experiments were conducted on a 15-inch MacBook Air (2024) with the following specifications:
\begin{itemize}
    \item \textbf{Chip:} Apple M3
    \item \textbf{Memory:} 24 GB
    \item \textbf{Operating System:} macOS 14.6.1 (23G93)
\end{itemize}

Since the computational heavy lifting was performed via the OpenAI API, the local machine's specifications did not significantly impact the performance of the experiments. Each API call, processing a single sample from the Tango dataset, took approximately 2 seconds. Processing 100 samples through our multi-agent system and the baseline GPT-4 model required a total of approximately 6.7 minutes per model, amounting to 13.4 minutes in total.

The overall compute resources required for the project were minimal, and there were no substantial computational constraints. No additional compute resources were utilized beyond what is reported here. Preliminary experiments and failed runs did not require significant additional compute time.

\end{document}